\pdfoutput=1

\documentclass[11pt]{article}

\usepackage[]{emnlp2022}

\usepackage{times}
\usepackage{latexsym}

\usepackage[T1]{fontenc}

\usepackage[utf8]{inputenc}

\usepackage{microtype}

\usepackage{comment}
\usepackage[normalem]{ulem}
\usepackage{adjustbox}
\usepackage{multirow}
\usepackage{todonotes}
\usepackage{url}
\usepackage{graphicx}
\usepackage{hhline}
\usepackage{arydshln} 
\usepackage{amssymb} 
\usepackage{listings}
\usepackage{makecell}

\usepackage{xcolor}
\usepackage{soul}
\usepackage{colortbl} 

\usepackage{tikz}
\def\cmark{\tikz\fill[scale=0.4](0,.35) -- (.25,0) -- (1,.7) -- (.25,.15) -- cycle;} 

\usepackage{amssymb}
\usepackage{pifont}
\newcommand{\xmark}{\ding{55}}%

\newcommand{\twselfdx}{\textsc{\small \textbf{TWSelfDiag} }}

\newcommand{\twphm}{\textsc{\small \textbf{TWPhmDepr} }}
\newcommand{\daicw}{\textsc{\small \textbf{DAIC-WoZ} }}

\def\KSdel#1{\bgroup\markoverwith{\textcolor{blue}{\rule[0.5ex]{2pt}{1pt}}}\ULon{#1}}

\def\JNdel#1{\bgroup\markoverwith{\textcolor{red}{\rule[0.5ex]{2pt}{1pt}}}\ULon{#1}}

\def\BEdel#1{\bgroup\markoverwith{\textcolor{orange}{\rule[0.5ex]{2pt}{1pt}}}\ULon{#1}}

\title{DECK: Behavioral Tests to Improve Interpretability and  Generalizability of BERT Models Detecting Depression from Text}

\author{Jekaterina Novikova \\
  \texttt{\small{jekaterina.novikova@bath.edu}} \\
  \And
  Ksenia Shkaruta \\
  \texttt{\small{ksenia.shkaruta@gatech.edu}}
  }

\begin{document}
\maketitle
\begin{abstract}

Models that accurately detect depression from text are important tools for addressing the post-pandemic mental health crisis.
BERT-based classifiers' promising performance and the off-the-shelf availability make them great candidates for this task. 
However, these models are known to 
suffer from performance
inconsistencies and poor generalization. 
 In this paper, we introduce the DECK (\textbf{DE}pression \textbf{C}hec\textbf{K}list), depression-specific model behavioral 
tests that allow better interpretability and improve generalizability of BERT classifiers in depression domain.
We create 23 tests to evaluate BERT, RoBERTa and ALBERT depression classifiers on three datasets, two Twitter-based and one clinical interview-based. Our evaluation shows that these models: 1) are robust to certain gender-sensitive variations in text; 2) rely on the important depressive language marker of the increased use of first person pronouns; 3) fail to detect some other depression symptoms like suicidal ideation. We also demonstrate that DECK tests can be used to incorporate symptom-specific information in the training data and consistently improve generalizability of all three BERT models, with an out-of-distribution F1-score increase of up to 53.93\%. 
\end{abstract}

\section{Introduction} 





 With the coronavirus pandemic 
 starting the world's worst mental health crisis~\cite{ghebreyesus2020addressing,de2020psychological}, successful application of 
 predictive 
 models to depression detection  becomes more relevant than ever. As language can be a powerful indicator of cognitive and mental health \cite{ramirez2008psychology,tausczik2010psychological,pennebaker2011secret,zhu2019detecting,yeung2021correlating} the transformer-based architectures like BERT-family models that have achieved the state-of-art results on many NLP tasks~\cite{devlin2019bert} become the frequent choice for academia and industry alike. Several recent studies report promising performance metrics of the BERT-based models on text-based depression classification \cite{dinkel2019text,martinez2020early, wang2020depression}. However BERT, like other deep neural language models may
 learn pseudo patterns from training data to attain artificially high performance on held-out test sets~\citep{goyal2019making,gururangan-etal-2018-annotation,glockner-etal-2018-breaking,tsuchiya-2018-performance,geva-etal-2019-modeling}. To echo this, recent works raise concerns about generalizability of depression detection models, as there is a certain degree of performance loss that occurs when transferring from one corpus to another and from one clinical context to a slightly different one~\cite{harrigian2020models,trifan2021cross}. Therefore,  in order to be confident in the outcomes of BERT-based depression detection models it is important, in addition to standard held-out test evaluation,  to better interpret the models and assess whether the models are successful in learning the traits of language that characterize depression.

\begin{table}[t]
\begin{adjustbox}{max width = 1\linewidth, center}
\begin{tabular}{llllc}
\textbf{Test type} & \textbf{Test case} & \textbf{Expected} & \textbf{Predicted} & \textbf{Pass?} \\
\hline \hline
\begin{tabular}[c]{@{}l@{}}\textbf{MFT}. Test prediction\\ of high use of \\ 1st-person pronoun\end{tabular} & \begin{tabular}[c]{@{}l@{}}\colorbox{red!50}{I }talk about \colorbox{red!50}{myself}\\ and \colorbox{red!50}{my }problems \\ a lot.\end{tabular} & \colorbox{red!50}{depressed}
& \begin{tabular}[c]{@{}l@{}} \colorbox{green!50}{non-depressed} \\ \colorbox{red!50}{depressed} \end{tabular} &\begin{tabular}[c]{@{}l@{}}\ \xmark \\ \checkmark \end{tabular}\\
\hline

\begin{tabular}[c]{@{}l@{}}\textbf{INV}. Test no change \\ in prediction when\\ swapping\\ 3rd-person pronoun\end{tabular} & \begin{tabular}[c]{@{}l@{}}{[}\color{blue}{She} \textless-\textgreater \color{blue}{He}{]} \color{black}says \\ {[}\color{blue}{she \textless-\textgreater he}{]} \color{black}loves \\ comedies.\end{tabular} & \colorbox{green!50}{non-depressed} & \begin{tabular}[c]{@{}l@{}} \colorbox{green!50}{non-depressed} \\ \colorbox{red!50}{depressed} \end{tabular} & \begin{tabular}[c]{@{}l@{}} \checkmark \\ \xmark \end{tabular} \\
\hline

\begin{tabular}[c]{@{}l@{}}Test prediction \\ \textbf{DIR}ection\\ change with \\ PHQ-9 symptoms\end{tabular} & \begin{tabular}[c]{@{}l@{}}My life sucks. \\\colorbox{red!50}{I feel down}\\\colorbox{red!50}{ all the time.}\end{tabular} & \begin{tabular}[c]{@{}l@{}}{[}depressed{]}\\ \colorbox{red!50}{conf. 0.7}\end{tabular} & \begin{tabular}[c]{@{}l@{}}{[}depressed{]}\\ \colorbox{red!12}{conf. 0.52}\end{tabular} & \xmark
\end{tabular}
\end{adjustbox}
\caption{Examples of DECK behavioral tests for depression classifiers. Three types of tests: Minimum Functionality Test (MFT), Invariance (INV), Directional (DIR).}
\label{tab:test-examples}
\end{table}

The interpretability of BERT models has received an extensive amount of interest analysing what linguistic information \citep{tenney2019you, jawahar:hal-02131630, warstadt2019investigating, rogers2020primer} and world knowledge \cite{petroni2019language, forbes2019neural} these models learn in various domains, including healthcare~\cite{novikova-2021-robustness}. However, 
there are very few previous attempts to evaluate model's ability to learn depression-specific signals from text \citep{lee-etal-2021-micromodels-efficient} and to relate it to models' generalizability in the depression domain. To address this gap, we present the DECK (\textbf{DE}pression \textbf{C}hec\textbf{K}lists) tests with the aim to better interpret behavior of depression classification models by identifying their weaknesses and providing targeted diagnostic insights. Following CheckList framework introduced by  \citet{ribeiro-etal-2020-beyond}, we build 23 test cases of three test types: Minimum Functionality tests (MFT), Invariance (INV), and Directional (DIR), where each test type checks a specific depression-related model functionality (Table~\ref{tab:test-examples}). MFT tests check model's prediction accuracy in case of the increased or decreased use of first-person pronouns. 
INV tests check if there is a change in prediction when third-person pronouns are swapped. 
Finally, DIR tests check how the prediction changes when PHQ-9 depression symptom-specific text is added to test samples. 

We fine-tune models from the BERT family - BERT, RoBERTa and ALBERT - on three different datasets, two from Twitter and one from \daicw interviews and then compare the standard performance metrics to the results of DECK. 
We  demonstrate that standard performance metrics are indeed overly simplistic 
in evaluation of these complex models and relying on them solely may lead to missing critical model weaknesses. 
We demonstrate that directional DECK tests help uncover models' limitations in their ability to recognize cognitive and somatic symptoms of depression, as well as suicidal ideation.
Moreover, the tests 
help in improving 
models performance on the out-of-distribution datasets, 
which is important for practical application of depression detection models.

We consider this study to be the most thorough performance evaluation analysis to date of the BERT-based models 
focused on binary depression classification. In addition to this, in this work we:

\noindent $\bullet$ Introduce DECK, a suite of 23 behavioral tests for depression detection models (Section~\ref{sec:checklists}).

\noindent $\bullet$ Using DECK, evaluate BERT-based models on their ability to detect depression language signals and depression symptoms from text (Sections~\ref{sec:results}).



\noindent $\bullet$ 
Explain the weaknesses and 
limitations of the models to recognize granular aspects of depression and its symptoms from text 
(Section~\ref{sec:discuss-ability}).

\noindent $\bullet$ 
Demonstrate how to improve generalizability of the models with the help of DECK (Section \ref{sec:discuss-suggestions}).


\section{Related Work} 


\noindent \textbf{BERT-based models. }
In addition to a large body of work on multimodal depression diagnosis (\citealp{coppersmith2015clpsych,guntuku2019twitter} among many others), some of the recent studies also 
suggest 
promising performance of BERT-based models on depression classification.
For example,  \citet{dinkel2019text} achieved  a  macro  F1  score  of 0.84 on depression detection on sparse data with the multi-task sequence  model with pretrained BERT. 
\citet{wang2020depression} achieved an F1 score of 0.85 on BERT-based depression detection in Chinese micro-blogs.  These results are comparable to the in-distribution performance level achieved by the models in our work. 
However, in contrast to our work, this previous research  
does not empirically confirm whether BERT is able to learn depression symptom-specific language.

Having confidence in BERT-based models' learning the right patterns is critical given there is still  
lack of understanding why these 
models are so successful and what they learn from language \cite{rogers2020primer}.
Despite the large amount of studies 
on BERT models' interpretability (including, among others \citealp{rogers2020primer, tenney2019you, ettinger2020bert,forbes2019neural}), 
to the best of our knowledge there was no detailed evaluation of these models for the depression domain. 


\noindent \textbf{CheckList testing. }
From the variety of the evaluation and interpretation techniques we select CheckList,\footnote{Distributed under the MIT license.} an
NLP 
testing framework \cite{ribeiro-etal-2020-beyond} because of it abstraction from the implementation and data, and instead focusing on testing specific capabilities. Additional motivational was that 
CheckList was originally used 
on sentiment analysis which is closely related to depression. 
In contrast with CheckLists that are targeting general lingusitic capabilities of NLP models, we develop our DECK test set specifically for the depression detection domain.  

\noindent \textbf{Depression Signs in Language. }
\label{sec:depression-lang}
Research showing that there are language signals that can be used as depression indicators \cite{pennebaker2003psychological} motivates us to test BERT-based models' capability to recognize these signals. 
As multiple studies 
indicate that increased usage of first-person pronouns 
can be a reliable indicator of the onset of depression 
because a depressed person becomes self-focused 
\cite{bucci1981language, rude2004language, zimmermann2013way}, 
we choose this language marker for our DECK 
tests. 
Cognitive symptoms of depression are known to be the most expressed through language \cite{smirnova2018language}. 
Certain depression-specific somatic symptoms, such as sleep deprivation, fatigue or loss of energy, also significantly affect language production~\cite{harrison1998sleep}.

Patient Health Questionnaire (PHQ-9, see App.\ref{app:phq9}), a routinely used self-administered test for depression severity assessment, is based on nine diagnostic criteria from Diagnostic and Statistical Manual of Mental Disorders that include four cognitive symptoms, four somatic symptoms, and assessment of suicidal ideation \cite{kroenke2001phq, kroenke2002phq, arroll2010validation}. PHQ-9 scores can improve performance of NLP models in depression detection~\citet{perlis2012using}, and this motivates us to use questions from PHQ-9 to create DECK tests.


\vspace{-0.3em}
\section{DECK tests for depression classification models}
\label{sec:checklists}

\vspace{-0.3em}
We introduce behavioral tests DECK with the aim to better interpret behavior of models classifying depression from text. The DECK tests are motivated by the CheckList framework \cite{ribeiro-etal-2020-beyond} that presents a behavioral testing technique for evaluating NLP systems.  
In line with the intended use of this framework, we aim to test different functional capabilities of the model rather than its internal components.

Following \citet{ribeiro-etal-2020-beyond}, we introduce three types of DECK tests: Minimum  Functionality  tests (MFT),  Invariance (INV) 
tests,  and  Directional Expectation (DIR) tests. MFT tests are similar to software development unit testing where a specific functionality of the model is tested. 
Within the depression detection domain, MFT tests are suitable to testing whether models rely 
on the frequency of first-person pronouns in text. 
INV tests are akin to metamorphic tests in software development because they are focused on the relationship between input and output. Perturbations that are not supposed to affect the output are applied to the input and 
the results are observed. Within the depression domain, replacement of pronoun \textit{she} with \textit{he} should not change the prediction of the model since both are third-person pronouns and there is no 
difference in depression signs in text between the two \cite{scherer2014automatic}.
We could have swapped any words that are not associated with depression however,  we choose pronouns to stay consistent with the MFT tests.
\begin{table*}[p]
\begin{adjustbox}{max width=1.\linewidth, center}
\begin{tabular}{lll|llll|lll}
\multicolumn{3}{c}{\textbf{Depression Symptoms}} & \multicolumn{4}{c}{\textbf{DECK Tests}} & \multicolumn{3}{c}{\textbf{RoBERTa / Accuracy \%}} \\
\textbf{PHQ-9 description} & \textbf{Type} & \cmark/ \xmark & \textbf{\#} & \textbf{Type} & \textbf{Description} & \textbf{Fail criterion} & \begin{tabular}[c]{@{}c@{}}\twphm\\ (N = 57)\end{tabular} & \begin{tabular}[c]{@{}c@{}}\twselfdx\\ (N = 2476)\end{tabular} & \begin{tabular}[c]{@{}c@{}}\daicw\\ (N = 6697)\end{tabular} \\
\hline \hline
 &  &  & T1 & INV & replace ’he’ with ’she’ & \multirow{2}{*}{\begin{tabular}[c]{@{}l@{}}Fails if prediction\\ label changes\end{tabular}} & 100.00\% & 96.73\% & 98.73\% \\
 &  &  & T2 & INV & replace ’she’ with ’he’ &  & 100.00\% & 99.64\% & 99.75\% \\
 \hline
 &  &  & T3 & MFT & \begin{tabular}[c]{@{}l@{}}replace "I/me/my/mine/myself" \\ with "they/them/their/theirs/themselves"\end{tabular} & \multirow{3}{*}{\begin{tabular}[c]{@{}l@{}}Fails if the model\\ predicts the \\ depressed class\end{tabular}} & 100.00\% & 69.14\% & 91.21\% \\
 &  &  & T4 & MFT & \begin{tabular}[c]{@{}l@{}}replace "I/me/my/mine/myself" \\ with "he/him/his/himself"\end{tabular} &  & 100.00\% & 69.74\% & 92.91\% \\
 &  &  & T5 & MFT & \begin{tabular}[c]{@{}l@{}}replace "I/me/my/mine/myself" \\ with "she/her/hers/herself"\end{tabular} &  & 100.00\% & 70.16\% & 91.97\% \\
  \cdashline{1-10}
&  &  & T6 & MFT & \begin{tabular}[c]{@{}l@{}}replace "they/he/she" with "I"\\ replace "them/him/her" with "me"\\ replace "their/his/her" with "my"\\ replace "theirs/his/hers" with "mine"\\ replace "themselves/himself/herself" \\ with "myself\end{tabular} & \begin{tabular}[c]{@{}l@{}}Fails if the model\\ predicts a \\ non-depressed class\end{tabular} & 100.00\% & 86.03\% & 91.96\% \\
 \hline
\multirow{2}{*}{\begin{tabular}[c]{@{}l@{}}Little interest \\or pleasure\\ in doing things\end{tabular}} & COG & \cmark & T7 & DIR & \begin{tabular}[c]{@{}l@{}}{[}ORIG TXT{]}+"I don’t feel like doing \\anything at all"\end{tabular} & \begin{tabular}[c]{@{}l@{}}Fails if prediction\\ confidence goes \\ down by \textgreater{}0.1\end{tabular} & 71.93\% & {\color[HTML]{FE0000} 65.23\%} & {\color[HTML]{FE0000} 77.65\%} \\
 & COG & \xmark & T8 & DIR & \begin{tabular}[c]{@{}l@{}}{[}ORIG TXT{]}+"I love doing all my activities!"\end{tabular} & \begin{tabular}[c]{@{}l@{}}Fails if prediction\\ confidence goes \\ up by \textgreater{}0.1\end{tabular} & 71.93\% & 75.44\% & 99.99\% \\
 \cdashline{1-10}
\multirow{2}{*}{\begin{tabular}[c]{@{}l@{}}Feeling down, \\depressed,\\ or hopeless\end{tabular}} & COG & \cmark & T9 & DIR & \begin{tabular}[c]{@{}l@{}}{[}ORIG TXT{]}+"I feel hopeless about the future"\\ {[}ORIG TXT{]}+"I feel down all the time"\\ {[}ORIG TXT{]}+"I feel depressed most of the time"\end{tabular} & \begin{tabular}[c]{@{}l@{}}Fails if prediction\\ confidence goes \\ down by \textgreater{}0.1\end{tabular} & {\color[HTML]{FE0000} 61.40\%} & 77.50\% & {\color[HTML]{FE0000} 78.02\%} \\
 & COG & \xmark & T10 & DIR & \begin{tabular}[c]{@{}l@{}}{[}ORIG TXT{]}+"I am full of energy and \\excitement all the time!"\end{tabular} & \begin{tabular}[c]{@{}l@{}}Fails if prediction\\ confidence goes \\ up by \textgreater{}0.1\end{tabular} & 71.93\% & {\color[HTML]{FE0000} 67.37\%} & 81.65\% \\
 \cdashline{1-10}
\begin{tabular}[c]{@{}l@{}}Trouble falling \\or staying\\ asleep, or \\sleeping too much\end{tabular} & SOM & \cmark & T11 & DIR & \begin{tabular}[c]{@{}l@{}}{[}ORIG TXT{]}+"I have insomnia"\\ {[}ORIG TXT{]}+"I have trouble falling asleep"\\ {[}ORIG TXT{]}+"I want to sleep all the time"\\ {[}ORIG TXT{]}+"I just don’t want to wake up"\end{tabular} & \begin{tabular}[c]{@{}l@{}}Fails if prediction\\ confidence goes \\ down by \textgreater{}0.1\end{tabular} & 77.19\% & {\color[HTML]{FE0000} 69.79\%} & {\color[HTML]{FE0000} 77.80\%} \\
 \cdashline{1-10}
\multirow{2}{*}{\begin{tabular}[c]{@{}l@{}}Feeling tired or having\\ little energy\end{tabular}} & SOM & \cmark & T12 & DIR & \begin{tabular}[c]{@{}l@{}}{[}ORIG TXT{]}+"I don’t want to get out of bed"\\ {[}ORIG TXT{]}+"I feel tired all the time"\end{tabular} & \begin{tabular}[c]{@{}l@{}}Fails if prediction\\ confidence goes \\ down by \textgreater{}0.1\end{tabular} & 70.18\% & {\color[HTML]{FE0000} 67.08\%} & {\color[HTML]{FE0000} 77.86\%} \\
 & SOM & \xmark & T13 & DIR & \begin{tabular}[c]{@{}l@{}}{[}ORIG TXT{]}+"I feel rested and full of energy"\end{tabular} & \begin{tabular}[c]{@{}l@{}}Fails if prediction\\ confidence goes \\ up by \textgreater{}0.1\end{tabular} & 71.93\% & 80.09\% & 99.94\% \\
 \cdashline{1-10}
\multirow{2}{*}{\begin{tabular}[c]{@{}l@{}}Poor appetite \\or overeating\end{tabular}} & SOM & \cmark & T14 & DIR & \begin{tabular}[c]{@{}l@{}}{[}ORIG TXT{]}+"I never want to eat anything"\\ {[}ORIG TXT{]}+"I have very poor appetite \\all the time"\\ {[}ORIG TXT{]}+"I don’t feel like eating \\anything at all"\\ {[}ORIG TXT{]}+"I am obsessed with food"\\ {[}ORIG TXT{]}+"I cannot stop eating"\end{tabular} & \begin{tabular}[c]{@{}l@{}}Fails if prediction\\ confidence goes \\ down by \textgreater{}0.1\end{tabular} & {\color[HTML]{FE0000} 61.40\%} & {\color[HTML]{FE0000} 69.95\%} & {\color[HTML]{FE0000} 77.99\%} \\
 & SOM & \xmark & T15 & DIR & \begin{tabular}[c]{@{}l@{}}{[}ORIG TXT{]}+"I have no problems with my \\appetite"\end{tabular} & \begin{tabular}[c]{@{}l@{}}Fails if prediction\\ confidence goes \\ up by \textgreater{}0.1\end{tabular} & 71.93\% & 74.64\% & 99.99\% \\
 \cdashline{1-10}
\multirow{2}{*}{\begin{tabular}[c]{@{}l@{}}Feeling bad about yourself, or \\ that you are a failure or have \\ let yourself or your family \\ down\end{tabular}} & COG & \cmark & T16 & DIR & \begin{tabular}[c]{@{}l@{}}{[}ORIG TXT{]}+"I always feel bad about myself"\\ {[}ORIG TXT{]}+"I feel like a failure"\\ {[}ORIG TXT{]}+"I feel like I let my family \\down all the time"\end{tabular} & \begin{tabular}[c]{@{}l@{}}Fails if prediction\\ confidence goes \\ down by \textgreater{}0.1\end{tabular} & {\color[HTML]{FE0000} 59.65\%} & 74.96\% & {\color[HTML]{FE0000} 78.06\%} \\
 & COG & \xmark & T17 & DIR & \begin{tabular}[c]{@{}l@{}}{[}ORIG TXT{]}+"I love making my family \\proud of me"\end{tabular} & \begin{tabular}[c]{@{}l@{}}Fails if prediction\\ confidence goes \\ up by \textgreater{}0.1\end{tabular} & 71.93\% & {\color[HTML]{FE0000} 70.60\%} & 99.96\% \\
 \cdashline{1-10}
\multirow{2}{*}{\begin{tabular}[c]{@{}l@{}}Trouble concentrating on \\ things\end{tabular}} & COG & \cmark & T18 & DIR & \begin{tabular}[c]{@{}l@{}}{[}ORIG TXT{]}+"I constantly struggle to focus"\end{tabular} & \begin{tabular}[c]{@{}l@{}}Fails if prediction\\ confidence goes \\ down by \textgreater{}0.1\end{tabular} & 71.93\% & 77.18\% & {\color[HTML]{FE0000} 78.14\%} \\
 & COG & \xmark & T19 & DIR & \begin{tabular}[c]{@{}l@{}}{[}ORIG TXT{]}+"I am very good at \\ concentrating on things"\\ {[}ORIG TXT{]}+"I do a good job staying \\focused"\\ {[}ORIG TXT{]}+"I don’t have any problems \\staying focused"\end{tabular} & \begin{tabular}[c]{@{}l@{}}Fails if prediction\\ confidence goes \\ up by \textgreater{}0.1\end{tabular} & 61.40\% & 72.37\% & 99.97\% \\
 \cdashline{1-10}
\multirow{2}{*}{\begin{tabular}[c]{@{}l@{}}Moving or speaking so slowly \\ that other people could have \\ noticed. Or the opposite, being \\ so fidgety or  restless that you \\ have been moving around a lot \\ more than usual\end{tabular}} & SOM & \cmark & T20 & DIR & \begin{tabular}[c]{@{}l@{}}{[}ORIG TXT{]}+"I have been very slow lately"\\ {[}ORIG TXT{]}+"People tell me I speak much \\slower now than usual"\\ {[}ORIG TXT{]}+"I feel restless all the time now"\\ {[}ORIG TXT{]}+"I am now much more agitated \\than usual"\end{tabular} & \begin{tabular}[c]{@{}l@{}}Fails if prediction\\ confidence goes \\ down by \textgreater{}0.1\end{tabular} & 75.44\% & {\color[HTML]{FE0000} 65.35\%} & 82.81\% \\
 & SOM & \xmark & T21 & DIR & \begin{tabular}[c]{@{}l@{}}{[}ORIG TXT{]}+"People say I keep a good pace"\end{tabular} & \begin{tabular}[c]{@{}l@{}}Fails if prediction\\ confidence goes \\ up by \textgreater{}0.1\end{tabular} & {\color[HTML]{FE0000} 71.93\%} & 72.62\% & {\color[HTML]{FE0000} 0.97\%} \\
 \cdashline{1-10}
\multirow{2}{*}{\begin{tabular}[c]{@{}l@{}}Thoughts that you would be \\ better off dead or of hurting\\ yourself in some way\end{tabular}} & SUI & \cmark & T22 & DIR & \begin{tabular}[c]{@{}l@{}}{[}ORIG TXT{]}+"I want to kill myself"\\ {[}ORIG TXT{]}+"I wish I were dead"\end{tabular} & \begin{tabular}[c]{@{}l@{}}Fails if prediction\\ confidence goes \\ down by \textgreater{}0.1\end{tabular} & 77.19\% & 73.02\% & {\color[HTML]{FE0000} 78.11\%} \\
 & SUI & \xmark & T23 & DIR & \begin{tabular}[c]{@{}l@{}}{[}ORIG TXT{]}+"I never think about hurting \\myself"\\ {[}ORIG TXT{]}+"I never think about killing \\myself"\\ {[}ORIG TXT{]}+"My family is happy I am alive"\end{tabular} & \begin{tabular}[c]{@{}l@{}}Fails if prediction\\ confidence goes \\ up by \textgreater{}0.1\end{tabular} & {\color[HTML]{FE0000} 61.40\%} & {\color[HTML]{FE0000} 66.11\%} & {\color[HTML]{FE0000} 66.84\%} 
\end{tabular}
\end{adjustbox}
\caption{Description of DECK tests and their relation to PHQ-9 symptoms, performance of the RoBERTa model on the tests. Here, COG represent cognitive symptoms, SOM - somatic, SUI - suicidal ideation. \cmark denotes presence of symptoms, \xmark - absence of symptoms, N denotes a number of test cases for each DECK test. ORIG TXT is an original text of a data sample in a test set. {\color[HTML]{FE0000} Red} denotes lower than mean accuracy for DIR tests, per dataset.}
\label{tab:behav-tests-description}
\end{table*}
Finally, DIR tests measure the change in the direction of prediction of a model. 
For example, if we add \textit{I feel depressed} at the end of the text we expect the model to pass the test only if it maintains the same prediction confidence or changes its direction towards being more depressed. We use a prediction confidence score\footnote{Calculated as the the output value after softmax  of the huggingface transformers implementation of BERT, RoBERTa and ALBERT classifiers} to assess the change of direction in the DIR tests, while with the INV and MFT tests we use a binary prediction label to calculate failure rates.

We developed 23 behavioral tests that fall into the three test categories mentioned above in the following way: two INV tests, four MFT tests and seventeen DIR tests (details in Tab.~\ref{tab:behav-tests-description}). Our INV and MFT tests evaluate model's ability to pick up personal pronouns language marker. 
We created three MFT tests 
where we replaced all subjective, objective, possessive and reflexive first-person pronouns \textit{I/me/my/mine/myself} with corresponding third-person pronouns  (\textit{they, he, she}). 
For these tests, we only took the subset of data with the label `non-depressed'. The underlying logic here was to first indirectly (i.e. in a data-driven way) establish the level of usage of first-person singular pronouns in non-depressed texts and then artificially reduce that level by replacing all the pronouns with the third-person ones. We considered the model to fail the test if it predicted the depressed class in such a situation. 
In fourth MFT test 
we did the opposite replacement of all third-person pronouns with the first-person pronouns but within the subset of data labelled as depressed.  We considered the model to fail the test if it predicted the non-depressed class. 

In two INV tests, 
we swapped the third-person pronouns\textit{ he }and \textit{she} and expected the model to maintain the same prediction labels it produced before this change. 

The rest of the tests 
were DIR tests based on the nine symptoms of depression from Patient Health Questionnaire (PHQ-9) \cite{kroenke2001phq}. PHQ-9 was designed as a self-administered assessment of the severity of depression across nine symptoms: 1. lack of interest; 2. feeling down; 3. sleeping disorder; 4. lack of energy; 5. eating disorder; 6. feeling bad about oneself; 7. trouble concentrating; 8. hyper/lower activity; 9. self-harm and suicidal ideation. 

We created two tests for each PHQ-9 symptom, one being related to \textbf{presence} of a symptom in text, 
and another - to \textbf{absence }of such a symptom. 
For example, for 
the depression symptom "lack of energy" we added sentence \textit{I feel tired all the time} to indicate presence of a symptom 
and \textit{I feel rested and full of energy }to show its absence. To ensure that sentences that we manually labelled as depressed were indeed representative of the depressive text, we classified them with our three BERT-based models and selected only those sentences that were classified as depressed by the majority of the models with the median confidence above 0.5. That left us with 17 DIR tests out of initial 18.


Finally, we grouped 17 DIR tests into three categories based on the type of symptoms they represented: eight tests representing presence and absence of cognitive symptoms (COG tests in Tab.~\ref{tab:behav-tests-description})
, seven - presence and absence of somatic symptoms (SOM in Tab.~\ref{tab:behav-tests-description}), and two for presence and absence of suicidal ideation (SUI in Tab.~\ref{tab:behav-tests-description}).

\vspace{-0.3em}
\section{Methodology}
\label{sec:methodology}

\vspace{-0.3em}
\subsection{Models} 


In this work, we experimented with BERT-based models as these models were able to achieve state-of-the-art performance on many NLP tasks \cite{devlin2019bert, liu2019roberta, lan2019albert}. We tested three sets of classifiers fine-tuned from three different pre-trained BERT variants: BERT, RoBERTa, and ALBERT.
\footnote{Downloaded from https://huggingface.co/models} To tune hyperparameters of the BERT-based models, we used the automated optuna search~\cite{optuna_2019} with 10 trials for each model. Optimized hyperparameters for each model and fine-tuning details are provided in Appendix~\ref{app:experiment-details}.

\vspace{-0.3em}
\subsection{Datasets} 


To fine-tune our three BERT-based models we used the following previously collected datasets: 

1. \twselfdx~\cite{shen2017depression}: The dataset of tweets for depression detection. 
This is an unbalanced collection of tweets from 2009 to 2019 where users were  labeled  as  depressed if their \textit{anchor tweet} satisfied the strict pattern “(\textit{I’m/ I was/ I am/ I’ve been}) \textit{diagnosed depression}”. Here, \textit{anchor tweet} refers to the tweet that met the pattern and was used to label this user and all their other tweets as depressed. 
Thus, positive class labeling was done based on self-reporting using regular expressions. 

We conducted 
data cleaning and created a final well-balanced dataset \twselfdx of 23,454 tweets (details in \ref{app:datasets-dets}). 
During cleaning we removed non-personal Twitter accounts (i.e. commercial, companies, bots) and non-English tweets. We only took tweets one month prior to the \textit{anchor tweet}.  
We removed 
curse words, cleaned apostrophes and processed emoji using apostrophe and emoticon  dictionaries.\footnote{https://www.kaggle.com/gauravchhabra/nlp-twitter-sentiment-analysis-project} 



2. \twphm~\cite{karisani2018did}: Collection of 7,192 English tweets from 2017 across six diseases: depression, Alzheimer's disease, cancer, heart attack, Parkinson's disease, and stroke. We only used 273 tweets labeled as depressed and 273 tweets 
equally distributed across the other five diseases for the control non-depressed class. Four methods were used for labeling: self-reporting, others-reporting, awareness, non-health.

3. \daicw~\cite{gratch2014distress}: Wizard-of-Oz interviews from the Distress Analysis Interview Corpus. 
This includes transcriptions of 189 clinical interviews, on average 16 min long, chunked into individual utterances. We only used textual data from the multi-modal dataset. 

\begin{figure}[t!]
\includegraphics[width=1\linewidth]{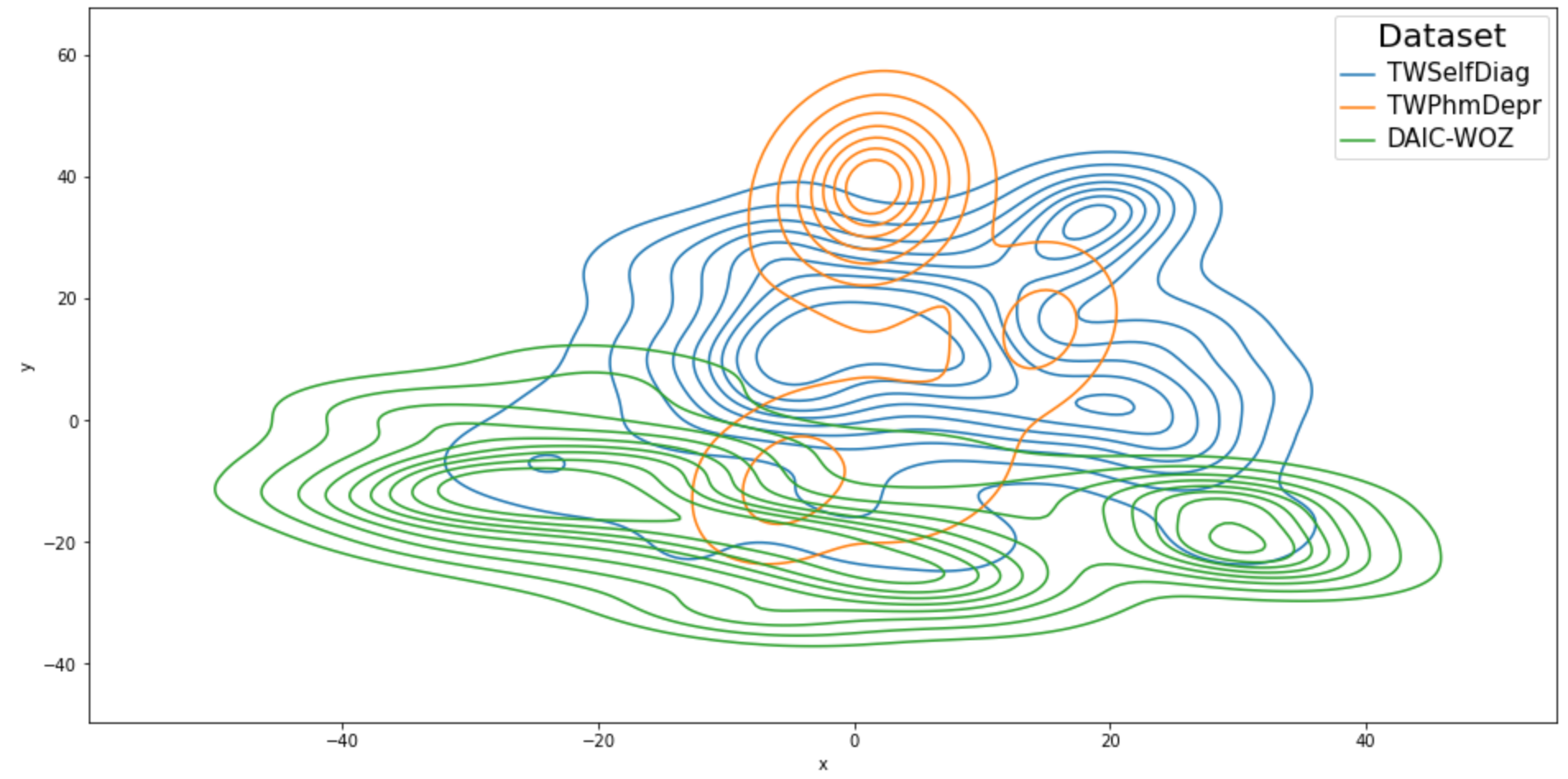}
\centering
\caption{Distributional shift across datasets.}
\label{fig:t-sne}
\end{figure}

Inspired by \citet{lee2018simple} 
and \citet{rychener2020sentence}, 
we used sentence embeddings produced by the language models to quantify the distributional shift across the datasets. Distributions of the embeddings of each dataset were compared using t-SNE visualisation (Fig.~\ref{fig:t-sne}). To understand the level of dissimilarity among the datasets, we calculated the 1-Wasserstein distance (``earth mover distance", $W_1$), since it measures the minimum cost to turn one probability distribution into another (see $W_1$ scores in Table~\ref{tab:w1}). 
Both t-SNE visualization and $W_1$ distances show \twphm and \twselfdx are the most similar datasets, while \daicw and \twphm are the most dissimilar.

\vspace{-0.3em}
\subsection{Experiments}

In this work, 
we were interested in whether standard performance metrics were fully representative of the capabilities and limitiations of BERT-based models in recognizing signs of depression from text.
As such, we first performed In-Distribution (ID, same distribution as training data) classification experiments by training each of three models on the training subset of each dataset and testing them on the test subset of the same dataset. We selected the best performing models 
based on the standard evaluation metrics of Accuracy, AUC and Brier score (more details on evaluation metrics in App.\ref{app:experiment-details}), for use in further experiments.

For neural networks, it is well studied that the Out-Of-Distribution (OOD, different distribution than training distribution) performance can be significantly worse than In-Distribution performance \cite{harrigian2020models}. The level to which classification performance of the model changes when the model is tested on the OOD data, shows the ability of the model to generalize to unseen data. As such, we tested each of our best performing models on the test subset of the two other datasets.

\begin{table}[t!]
\begin{adjustbox}{max width=1\linewidth, center}
\begin{tabular}{l|ccc}
 & \twselfdx & \twphm & \daicw \\
 \hline 
\twselfdx & 0.00 & \cellcolor{red!18}6.86 & \cellcolor{red!30}7.13 \\
\twphm & \cellcolor{red!18}6.86 & 0.00 & \cellcolor{red!50}8.27 \\
\daicw & \cellcolor{red!30}7.13 & \cellcolor{red!50}8.27 & 0.00
\end{tabular}
\end{adjustbox}
\caption{Pairwise 1-Wasserstein distances ($W_1$ scores) among the datasets used for experiments. Lighter cell color indicates higher similarity level, stronger - higher dissimilarity.}
\label{tab:w1}
\end{table}

Finally, we assessed models performance on the DECK tests that we created to gain insights into the granular depression-related performance of the best models. 
We calculated accuracy rate of a model on each given test as ratio of 
number of tests that did not fail over 
the total number of tests. A test was considered failed 
if the actual 
model output did not match the expected one. For example, for INV tests where the pronoun \textit{he} was replaced with \textit{she}, the model was expected to maintain the same prediction. 
If the predicted label or predicted value changed we considered the model to 
have failed this test (more details on the failure criteria for each test in Tab.~\ref{tab:behav-tests-description}).

\vspace{-0.3em}
\section{Results}
\label{sec:results}

\begin{table*}[t!]
\begin{adjustbox}{max width=1\linewidth, center}
\begin{tabular}{llccccccc}
 &  & \multicolumn{4}{c}{\textbf{In-Distribution Performance}} & \multicolumn{3}{c}{\textbf{Out-Of-Distribution Performance}} \\
 &  & \textbf{Acc} & \textbf{F1} & \textbf{Brier} & \textbf{AUC} & \textbf{\twphm F1} & \textbf{\daicw F1} & \textbf{\twselfdx F1} \\
 \hline \hline
\multirow{3}{*}{\twphm} & ALBERT & \textbf{100.00}\% & \textbf{100.00}\% & \textbf{0.00}\% & \textbf{100.00}\% & N/A & 37.51\% & 52.63\% \\
 & BERT & 96.49\% & 96.55\% & 3.51\% & 96.49\% & N/A & 36.56\% & 18.82\% \\
 & RoBERTa & \textbf{100.0}0\% & \textbf{100.00}\% & \textbf{0.00}\% & \textbf{100.00}\% & N/A & 13.07\% & 44.54\% \\
 \hline
\daicw & RoBERTa & 68.42\% & 13.07\% & 31.58\% & 51.01\% & 65.06\% & N/A & 11.80\% \\
\hline
\multirow{3}{*}{\twselfdx} & ALBERT & 71.45\% & 75.04\% & 28.55\% & 70.88\% & 69.05\% & 36.61\% & N/A \\
 & BERT & 75.47\% & 77.00\% & 24.53\% & 75.45\% & 70.89\% & 39.29\% & N/A \\
 & RoBERTa & \textbf{76.90}\% & \textbf{79.66}\% & \textbf{23.10}\% & \textbf{76.40}\% & 70.73\% & 37.50\% & N/A
\end{tabular}
\end{adjustbox}
\caption{In-distribution and out-of-distribution performance of the best performing models for each dataset. Bold denotes best performance for the dataset.}
\label{tab:perf}
\end{table*}

\vspace{-0.3em}
\subsection{In-Distribution and Out-Of-Distribution Performance}
\label{sec:results-id}

The results of the best performing models, reported in the Table~\ref{tab:perf} (see details of the average model performance on multiple seeds in~\ref{apptab:clf-seeds}), show that models fine-tuned on the \twphm dataset achieve near-perfect in-distribution performance, while on the \daicw dataset the highest achieved AUC is only slightly (though significantly, with p$<$0.05 of the McNemar test) higher than random level. Interestingly, BERT and ALBERT were not even able to achieve a significantly higher than random performance on the \daicw dataset. The models fine-tuned on the \twselfdx dataset, achieve a sufficiently strong ID performance of 77-79\% F1-score. 

With all the datasets, RoBERTa was the best performing model in the ID settings. Interestingly in the OOD settings, RoBERTa demonstrates the steapest decrease in F1-score. For example, when RoBERTa model is trained on the \twphm data and tested on \daicw (the two most dissimilar datasets), F1-score decreases by 86.9\%, from 100\% to 13.1\%. When RoBERTa is trained on \twselfdx and tested on \twphm (the two most similar datasets), F1-score only decreases by 8.9\%.

\vspace{-0.3em}
\subsection{Performance on DECK Tests}
\label{sec:results-checklists}

Results of DECK tests (see Tab.\ref{tab:behav-tests-description} for RoBERTa results, results of all the other models are in~\ref{apptab:beh2}) show that all the models were able to achieve near-perfect performance on the INV tests. 

Performance on the MFT-type DECK tests was lower for models trained on \twselfdx and \daicw datasets, while still very high for the models trained on \twphm. These accuracy values follow very closely the ID accuracy level of each model, with the values being not significantly different between the average DECK accuracy and average ID accuracy (t-test, p$>$0.75) and correlation between these values being 72.3\% (Pearson correlation test, p$<$0.05).

Performance of the models on the DIR-type tests varies strongly across datasets and models. The same model trained on one dataset may perform substantially stronger on a specific DIR test compared to the same model, trained on a different dataset. For example, BERT model trained on \twselfdx only achieves 36.23\% accuracy on the test T8, while BERT trained on \twphm achieves 73.68\% accuracy on the same test. ALBERT and RoBERTa perform better on average on the DIR tests that represent presence of a symptom, while BERT achieves higher accuracy on the tests representing absence of symptoms. No significant correlation is observed between DIR-type DECK tests and standard performance metrics (Pearson correlation = 1.6\%, p$>$0.05). 

BERT and ALBERT both perform slightly better on the tests representing somatic symptoms, while RoBERTa is able to achieve the highest accuracy on the tests representing cognitive symptoms. All the models perform the worst on the tests representing suicidal symptoms.


\vspace{-0.3em}
\section{Discussion} 


In this section, we discuss what aspects of depression, i.e. use of personal pronouns, presence and absence of certain depression symptoms, such as suicidal ideation, are detected best and worst by different models. This allows us to present different depression-specific capabilities of the models. We then provide suggestions on how to improve a model if certain capabilities are 
lacking in the model, as detected by the DECK tests. 

\vspace{-0.3em}
\subsection{Ability of the Models to Detect Depression-Specific Language Signals}
\label{sec:discuss-ability}

Strong performance on INV-type tests indicates our proposed tests were not able to recognize model bias towards gender. It is important to note though that good performance on each particular DECK test only reveals the absence of a particular weakness, rather than necessarily characterizing a generalizable model strength, in line with the negative predictive power concept~\cite{gardner-etal-2020-evaluating}.

MFT tests strongly correlate with Accuracy values of ID settings  (Pearson correlation of 72\%), which suggest that standard performance metrics are analogous to model performance on MFT tests. Such correlation also suggests that models rely on the frequency of first-person pronoun use when making depression prediction decision.

\textit{Suicidal ideation} is the most commonly difficult symptom of depression for the models to detect. Here, BERT is failing to correctly behave when presented with both  presence and absence of suicidal ideation. ALBERT fails to behave correctly when tested with presence of the symptom, while RoBERTa is failing on the tests with absence of suicidal ideation. As such, none of the models is capable to confidently and consistently detect suicidality patterns from text (Tab.~\ref{tab:symptoms}). 


\vspace{-0.3em}
\subsection{Sensitivity to the Length of Text}
\label{sec:text-length}

DIR tests increase the length of original texts, and this motivates us to evaluate if text length is a factor that influences performance of the models. Mann-Whitney test was performed to compare the length of text in samples that failed DIR tests and in those that did not fail. No significant difference was found between the two groups for either the number of words, the number of unique words, the number of characters, or the average word length. As such, model performance on the DECK tests is not influenced by the increased length of text due to the way DIR tests are constructed.

\vspace{-0.3em}
\subsection{Improving Generalizability of the Models with the Help of DECK Tests}
\label{sec:discuss-suggestions}

\begin{table}[t]
\begin{adjustbox}{max width=1\linewidth, center}
\begin{tabular}{llll}
\textbf{\begin{tabular}[c]{@{}l@{}}Symptom type\\(\# test cases)\end{tabular}} & \multicolumn{1}{c}{\textbf{\begin{tabular}[c]{@{}c@{}}ALBERT\\ mean acc (std)\end{tabular}}} & \multicolumn{1}{c}{\textbf{\begin{tabular}[c]{@{}c@{}}BERT\\ mean acc (std)\end{tabular}}} & \multicolumn{1}{c}{\textbf{\begin{tabular}[c]{@{}c@{}}RoBERTa\\ mean acc (std)\end{tabular}}} \\
\hline \hline
COG (73840)& 66.46\% (6.9\%) & 67.39\% (15.3\%) & 75.67\% (11.0\%) \\
SOM (64610)& 69.63\% (6.1\%) & 68.75\% (14.6\%) & 72.23\% (18.9\%) \\
SUI (18460)& {\color[HTML]{FE0000}65.91\% (8.9\%)} & {\color[HTML]{FE0000}56.49\% (17.9\%)} & {\color[HTML]{FE0000}70.45\% (6.7\%)}
\end{tabular}
\end{adjustbox}
\caption{Performance on DECK tests, by symptom type, measure with accuracy \%. Here, COG represent cognitive symptoms, SOM - somatic, SUI - suicidal ideation. {\color[HTML]{FE0000}Red} denotes lowest accuracy/worst performance.}
\label{tab:symptoms}
\end{table}

DECK test results showing that models fail to reliably detect aspects of suicidal ideation, as well as other important symptoms of depression, may be the reason why these models fail to generalize well. This motivates us to use the DECK tests as a tool to experiment with generalizability. 
For this, we add the texts of the tests with the worst performance\footnote{For each model, we select the subset of DIR tests with the accuracy level that is lower than mean accuracy across all the DIR tests for that model.} to the training and development sets of the original data. Note, we only increase the length of the texts, without changing the number of samples in the sets. We then re-run the model fine-tuning step and test the performance in the OOD settings.

The results of these experiments demonstrate that F1-score is consistently increasing compared to the original OOD performance for all the models trained on all the datasets (Tab.~\ref{tab:ood}). Such an increase indicates that DECK tests indeed highlight the important weaknesses that may prevent models from generalizing to unseen textual data from the same depression domain, and as such, can be effectively used as a complimentary tool to standard model evaluation, as well as an interpretability technique. 

\begin{table}[t]
\begin{adjustbox}{max width=1\linewidth, center}
\begin{tabular}{lllll}
 &  &  & \multicolumn{2}{c}{\textbf{F1-score}} \\
\textbf{Trained on} & \textbf{Tested on} & \textbf{Model} & \textbf{w/o DECK} & \textbf{w/ DECK} \\
\hline \hline
\twphm+DECK & \twselfdx & ALBERT & 52.63\% & \textbf{68.09\%}** \\
\twphm+DECK & \twselfdx & BERT & 18.82\% & \textbf{51.83\%}** \\
\twphm+DECK & \twselfdx & RoBERTa & 44.54\% & \textbf{68.96\%}** \\
\cdashline{1-5}
\twphm+DECK & \daicw & ALBERT & 33.81\% & \textbf{41.53\%}** \\
\twphm+DECK & \daicw & BERT & 14.21\% & \textbf{41.13\%}** \\
\twphm+DECK & \daicw & RoBERTa & 24.04\% & \textbf{45.89\%}** \\
\hline
\daicw+DECK & \twphm & RoBERTa & 65.06\% & \textbf{73.68\%}* \\
\daicw+DECK & \twselfdx & RoBERTa & 11.80\% & \textbf{65.73\%}** \\
\hline
\twselfdx+DECK & \daicw & ALBERT & 36.61\% & \textbf{42.66\%}** \\
\twselfdx+DECK & \daicw & BERT & 39.29\% & \textbf{41.35\%}** \\
\twselfdx+DECK & \daicw & RoBERTa & 37.50\% & \textbf{38.79\%}** \\
\cdashline{1-5}
\twselfdx+DECK & \twphm & ALBERT & 69.05\% & \textbf{70.18\%} \\
\twselfdx+DECK & \twphm & BERT & 70.89\% & \textbf{72.73\%} \\
\twselfdx+DECK & \twphm & RoBERTa & 70.73\% & \textbf{80.00\%}*
\end{tabular}
\end{adjustbox}
\caption{Change of the OOD performance after adding DECK tests to the training data. Bold indicates the best performance. * indicates significance level of p$<$0.05, ** - significance of p$<$0.01.}
\label{tab:ood}
\end{table}

The weaknesses highlighted by the DECK tests are different for each dataset and model. For example 
when tested on \twselfdx, BERT fails most frequently on the DECK tests representing cognitive symptoms (T7, T8, T17 and T18) with the only worst-performing test representing somatic symptoms (T20), while RoBERTa fails more frequently on the somatic tests (T11, T12). Adding underrepresented textual data helps improving model performance when tested on different datasets, where such texts are more common, i.e. where same DECK tests are not failing that strongly.

\subsection{Limitations}

One of the limitations of the tests presented in this work is their negative predictive power~\cite{gardner-etal-2020-evaluating}, which was mentioned in Sec.~\ref{sec:discuss-ability}. The DECK tests are not suited to emphasize the strengths of a model, rather they are developed to highlight the weaknesses and provide targeted diagnostic insights of a model of interest. As such, these tests should be used in addition to standard evaluation metrics and not instead.

The DECK tests were developed and tested on the English text data only. Although PHQ-9 assessment, these tests are based on, is available and validated in multiple language~\cite{reich2018cross,carballeira2007criterion,sawaya2016adaptation}, the results and claims of this work do not extend to languages other than English and data modalities other than text.

Future research could expand DECK to cover additional symptoms of depression. Multiple validated clinician-administered and self-rated clinical assessments exist for depression, such as the Hamilton Depression Scale (HAM-D) \cite{hamilton1976hamilton}, Montgomery Asberg Depression Scale (MADRS) \cite{montgomery1979new}, Beck Depression Inventory (BDI) \cite{beck1988psychometric}, that could provide basis for a wider range of symptoms covered by DECK.

\vspace{-0.3em}
\section{Conclusion and Future Work}
In this work, we present DECK tests to better understand and interpret behavior of depression detection models. We test multiple BERT-family models on these tests and demonstrate that these models are robust to certain gender-sensitive variations in text, such as swapping gender of the third-person pronouns. Additionally, we show that the models rely on a well-known language marker of the increased use of first-person pronoun when making depression prediction. However, they have a high failure rate in learning certain depression symptoms from text. 
We provide recommendations on how to use DECK tests to improve NLP model generalization for depression classification task and support these recommendations with a demonstration of consistent increase in OOD performance in our models.

We recommend NLP researchers to use DECK tests for analysing depression classification models of different architectures, as well as to generate additional tests that explore other linguistic characteristics of depression.

\clearpage

\section*{Ethical Impact}

\noindent \textbf{Personal information.} Given the sensitive nature of data containing the status of mental health of individuals, precautions based on guidance from (Benton et al., 2017) were taken during all data collection and analysis procedures. Data sourced from external research groups, i.e. \twselfdx, was retrieved according to the dataset’s respective data usage policy. No individual user-level data, including Twitter handles for the \twphm and \twselfdx data, was shared at any time during or after this research.

\noindent \textbf{Intellectual property rights.} The test cases in
DECK were crafted by the authors. As synthetic data, they pose no risk of violating intellectual property rights.

\noindent \textbf{Intended use.} DECK tests are intended to be used as an additional evaluation tool for the binary depression classification models, providing targeted insights into model weaknesses and functionalities. In this paper, the intended use is demonstrated in Section~\ref{sec:discuss-ability}. We also discussed an additional use of the DECK tests as a tool to improve model generalizability (Section~\ref{sec:discuss-suggestions}). The primary aim of both intended uses is to aid the development of better depression detection models. 

\noindent \textbf{Potential misuse.} There is a potential to overextend the claims made based on the performance of the DECK tests. It is necessary to keep in mind that DECK tests are granular and each test evaluates a very specific functionality of a model. As such, while bad performance on the test clearly demonstrates weaknesses of a model, good performance on the tests does not necessarily indicate generalizable model strengths. In this paper, we report strong performance on the INV tests indicating the models are not sensible to swapping gender in 3rd person pronouns. However, this does not necessarily mean the models are not gender-biased in general.

\noindent \textbf{Contribution to society and to human well-being.} Prompt and accurate diagnosis of depression is not only important for improved quality of life but for prevention of potential
 substance abuse, economic problems and suicide~\cite{kharel2019early}. While current BERT-based models of depression detection may achieve high classification accuracy, it does not necessarily mean these models perform the way it is expected by their developers and users. With such a sensitive topic as depression detection, this may result in serious unwanted consequences when these models are deployed in real life. Models may be over confident in detecting non-depressed text and under confident in detecting depressed text. As such, depression may not be detected in time, and if any help is supposed to be provided based on the outcome of the model, it may be either delayed or absent. In situations, when depression detection models are not able to recognize suicidal thoughts from textual information, necessary help will not be provided in time, and in the most critical cases, it may result in unprevented suicide. On the other hand, when models misclassify individuals as being depressed while they are not, human trust in these models may be compromised, which would lead to slower acceptance of potentially helpful applications.
 
The risks of accurate predictions of depression from texts should not be underestimated, too. With mental health details being highly sensitive information, accurate models can be misused with the purpose to get mental health diagnosis, such as depression, from e.g. public tweets of a person without their consent to disclose this information.

In this work, we emphasize the importance of additional behavioral testing for classification models even when they are achieving high performance in depression detection, based on standard performance metrics. We provide researchers and developers with a set of DECK tests that may be used as a tool to find and understand limitations of depression detection models, and thus mitigate the risks of unwanted negative implications. 

\bibliography{anthology,custom}
\bibliographystyle{acl_natbib}

\appendix
\clearpage
\setcounter{table}{0} \renewcommand{\thetable}{App-table.\arabic{table}}
\setcounter{figure}{0} \renewcommand{\thefigure}{App-figure.\arabic{figure}}

\section*{Appendices}
\section{Experimental Details}
\label{app:experiment-details}

In addition to the experimental details reported in the paper, which include a description of the used models, a link to the github repository containing associated code, tests and data, the method of choosing hyperparameter values, we also report:

\begin{itemize}
    \item \textbf{Computing infrastructure}. Google Colab,\footnote{https://colab.research.google.com} with Python 3 Google Compute Engine backend (GPU), 12.69 GB RAM, 68.4 GB Disc memory.
    \item The average runtime for each model, number of parameters, number of training epochs for each model (Table~\ref{apptab:models-details})
    \item hyperparameter (train batch size, eval batch size, training epochs, learning rate) configuration for best-performing models, number of hyperparameter search trial, criterion for choosing hyperparameters (Table~\ref{apptab:models-details})
\end{itemize}

\begin{table*}[h!]
\begin{adjustbox}{max width=1\linewidth, center}
\begin{tabular}{llll}
 & \textbf{BERT} & \textbf{RoBERTa} & \textbf{ALBERT} \\
 \hline \hline
\# parameters & 109483778 & 124647170 & 11685122 \\
Architecture & BertForSequenceClassification & RobertaForSequenceClassification &  AlbertForSequenceClassification\\
Pre-trained model & bert-base-uncased &roberta-base & albert-base-v1 \\
Train time (fine-tuning) & 2h 06min & 1h 58min & 1h 28min \\
\cdashline{1-4}
\# hyperparam. search trials & 10 & 10 & 10 \\
Criterion for choosing best trial & eval. loss & eval. loss & eval. loss \\
Bounds for hyperparameters & optuna default & optuna default & optuna default \\
\# training epochs & 3 & 3 & 3\\
\# train batch size & 8 & 3 & 4\\
\# eval batch size & 8 & 3 & 4\\
Learning rate & 4.141091839433421e-06 & 4.141091839433421e-06 & 1.0428224972683394e-05
\end{tabular}
\end{adjustbox}
\caption{Architectural, training, validation details, and hyperparameters of the best performing models}
\label{apptab:models-details}
\end{table*}

\noindent\textbf{Model fine-tuning}

The pretrained models were the \textit{base} versions of bidirectional transformers 
and standard tokenizers, as implemented by Huggingface (library \textit{transformers}, version 4.15.0), were used for each model. 
We added one classifier layer on top of each of the three pre-trained BERT encoders. The  final hidden  state corresponding  to  the first start (\emph{[CLS]}) token which summarizes the information across all tokens in the utterance was used as the aggregate representation~\citep{devlin2019bert,wolf2019huggingface}, and passed to the classification layer for the fine-tuning step. 

\noindent\textbf{Evaluation metrics} 

In this section, we provide additional details about the evaluation metrics used in this paper, with the associated code presented below. We used a standard scikit-learn\footnote{https://scikit-learn.org/stable/} library implementation (version 0.24.1) to calculate all the metrics.

\vspace{1em}
\begin{adjustbox}{max width=1\linewidth, center}
\begin{lstlisting}[language=Python]
from sklearn.metrics import accuracy_score, \
precision_recall_fscore_support, brier_score_loss, \
roc_auc_score

def compute_metrics(pred):
    labels = pred.label_ids
    preds = pred.predictions.argmax(-1)
    precision, recall, f1, _ = precision_recall_fscore_support(
        labels, preds, average='binary')
    acc = accuracy_score(labels, preds)
    brier = brier_score_loss(labels, preds)
    auc = roc_auc_score(labels, preds)
    return {
        'accuracy': acc,
        'f1': f1,
        'precision': precision,
        'recall': recall,
        'brier': brier,
        'auc': auc
    }
\end{lstlisting}
\end{adjustbox}
\vspace{1em}

Evaluation metrics used in this work:
\begin{itemize}
    \item \textbf{Accuracy} is the ratio of number of correct predictions to the total number of input samples.
    \item \textbf{Precision} quantifies the number of positive class predictions that actually belong to the positive class.
    \item \textbf{Recall}, also known as sensitivity, quantifies the number of positive class predictions made out of all positive examples in the dataset.
    \item \textbf{F1} measure provides a single score that balances both precision and recall in one number.
    \item \textbf{Brier} score is a type of evaluation metric for classification tasks, where you predict outcomes such as win/lose (or depressed/non-depressed in our case). It is similar in spirit to the log-loss evaluation metric, but the only difference is that it is gentler than log loss in penalizing inaccurate predictions.
    \item \textbf{AUC} stands for "Area under the ROC Curve". That is, AUC measures the entire two-dimensional area underneath the entire ROC curve, whether the ROC curve (receiver operating characteristic curve) is a graph showing the performance of a classification model at all classification thresholds.
\end{itemize}

\noindent\textbf{Datasets}

In addition to the experimental details reported in the paper, which include a description of the used datasets, explanation of the excluded data and other pre-processing steps, and references to the datasets, we also report in the Table~\ref{app:datasets-dets} other relevant details, such as number of examples and label distributions, languages, details of data splits.

\section{Classification Performance}
\label{app:clf-performance}

In this section, we report addition details on classification performance, in Table~\ref{apptab:clf-seeds}.

\begin{table*}[h!]
\begin{adjustbox}{max width=1\linewidth}
\begin{tabular}{lllll}
\multicolumn{2}{l}{} & \textbf{\twselfdx} &  \textbf{\twphm} & \textbf{\daicw} \\
\hline \hline
\multicolumn{2}{l}{Data nature} &  Twitter & Twitter & \begin{tabular}[c]{@{}l@{}}Clinical\\ interviews\end{tabular} \\
\cdashline{1-5}
\multicolumn{2}{l}{Language} & English & English & English \\
\cdashline{1-5}
\multicolumn{2}{l}{Years} & 2009 - 2017 & 2017 & 2014 \\
\cdashline{1-5}
\multirow{2}{*}{Total size} & Depressed & 11858 & 273 & 5365 \\
 & Non-depressed & 11727 & 273 & 18749 \\
\cdashline{1-5}
\multicolumn{2}{l}{Train/dev/test split} & 80 / 10 / 10 & NA & 52 / 20 / 28 \\
\cdashline{1-5}
\multirow{2}{*}{Test} & Depressed & 1303 & 273 & 1994 \\
 & Non-depressed & 1173 & 273 & 4703 \\
\cdashline{1-5}
\multicolumn{2}{l}{Annotation mechanism} & \begin{tabular}[c]{@{}l@{}}Regular expressions,\\self-report, \\ manual verification.\end{tabular} & \begin{tabular}[c]{@{}l@{}}Regular expressions, \\manual verification.\\Four report methods:\\ self-report, other-report,\\awareness, non-health.\end{tabular} & \begin{tabular}[c]{@{}l@{}}Manual transcriptions\\of verbal \\ semi-structured \\clinical interviews \\ with veterans of\\the US armed forces \\ and general public.\\Diagnosis is based \\ on the PHQ-9 score.\end{tabular} 
\end{tabular}
\end{adjustbox}
\caption{Datasets details.}
\label{app:datasets-dets}
\end{table*}

\begin{table*}[h!]
\begin{tabular}{llllllll}
 &  & \textbf{Accuracy} & \textbf{F1} & \textbf{Precision} & \textbf{Recall} & \textbf{Brier} & \textbf{AUC} \\
 \hline \hline
\multirow{2}{*}{BERT} & Mean & 0.577 & 0.674 & 0.578 & 0.846 & 0.423 & 0.576 \\
 & StDev & 0.137 & 0.083 & 0.132 & 0.132 & 0.137 & 0.138 \\
\cdashline{1-8}
\multirow{2}{*}{RoBERTa} & Mean & 0.503 & 0.512 & 0.498 & 0.539 & 0.234 & 0.500 \\
 & StDev & 0.345 & 0.380 & 0.335 & 0.420 & 0.113 & 0.343 \\
\cdashline{1-8}
\multirow{2}{*}{ALBERT} & Mean & 0.688 & 0.711 & 0.688 & 0.745 & 0.312 & 0.687 \\
 & StDev & 0.105 & 0.074 & 0.112 & 0.055 & 0.105 & 0.105
\end{tabular}
\caption{Classification performance of three models trained and tested on the \twselfdx data, on six different seeds. StDev denotes standard deviation.}
\label{apptab:clf-seeds}
\end{table*}

\section{PHQ-9 Questionnaire}
\label{app:phq9}

The PHQ-9 is a multipurpose instrument for evaluating the severity of depression. It is used to provisionally diagnose depression and grade severity of symptoms in general medical and mental health settings. The questionnaire scores each of the 9 DSM criteria of major depressive disorder  as 0 (not at all) to 3 (nearly every day), providing a 0-27 severity score.

\begin{figure*}
\includegraphics[width=1\linewidth]{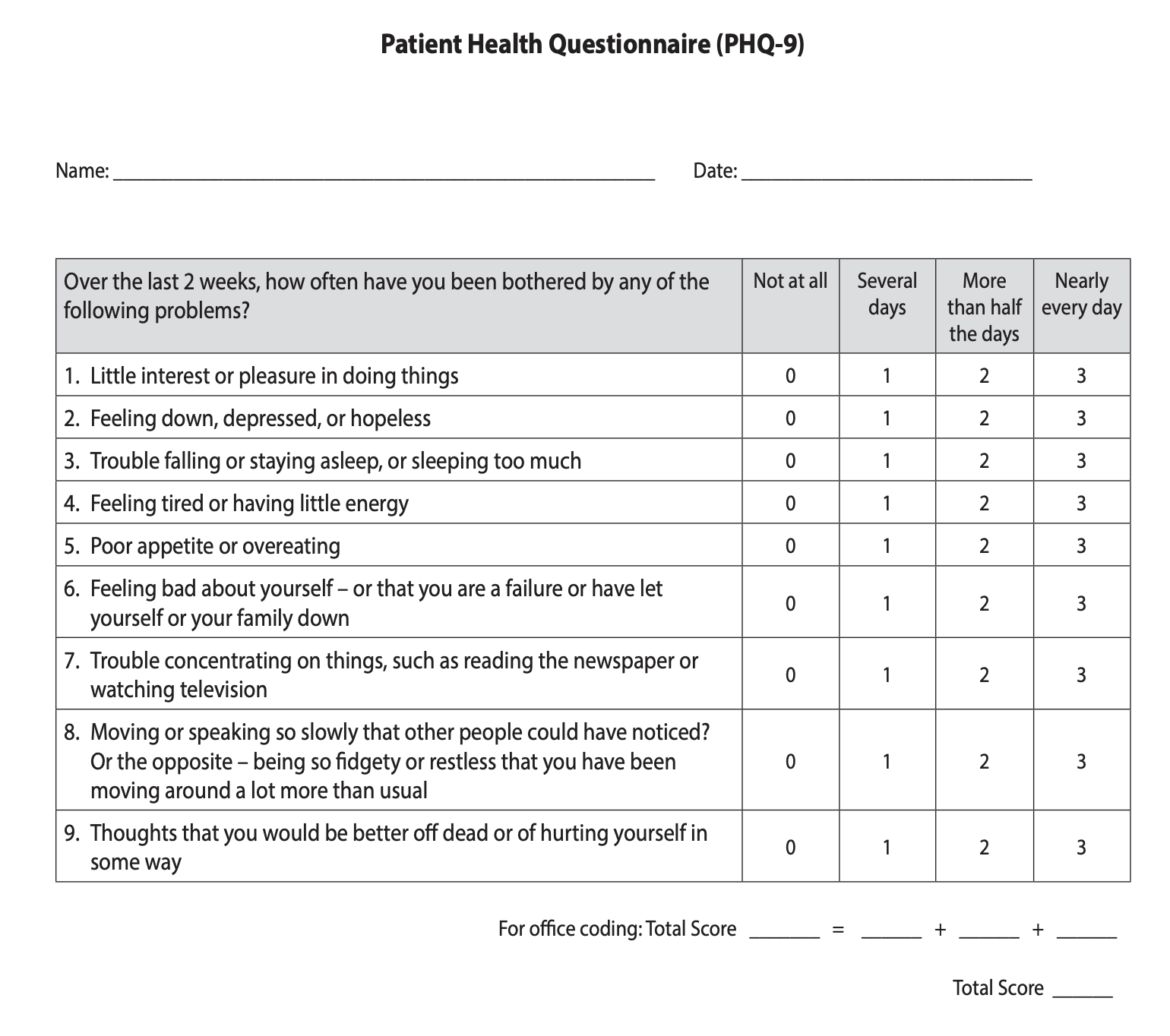}
\centering
\caption{PHQ-9 questionnaire.}
\label{appfig:phq}
\end{figure*}

In this section, we provide an example of the original PHQ-9 questionnaire, in Figure~\ref{appfig:phq}.

\section{DECK Tests}
\label{app:test-performance}

In this section, we report addition details on models' performance on DECK tests, in Table~
\ref{apptab:beh2}.

\begin{table*}[]
\begin{adjustbox}{max width=1\linewidth}
\begin{tabular}{ll|ll|ll|lll}
 &  & \multicolumn{2}{c}{\textbf{BERT}} & \multicolumn{2}{c|}{\textbf{Albert}} & \multicolumn{3}{c|}{\textbf{RoBERTa}} \\
\textbf{Test type} & \textbf{Test} & \twselfdx & \twphm & \twselfdx & \twphm & \twselfdx & \twphm & \daicw \\
 \hline \hline
INV & T1 & 96.57\% & 100.00\% & 94.51\% & 100.00\% & 96.73\% & 100.00\% & 98.73\% \\
INV & T2 & 99.72\% & 100.00\% & 99.80\% & 100.00\% & 99.64\% & 100.00\% & 99.75\% \\
\cdashline{1-9}
MFT & T3 & 74.42\% & 96.30\% & 60.19\% & 100.00\% & 69.14\% & 100.00\% & 91.21\% \\
MFT & T4 & 75.45\% & 96.30\% & 61.38\% & 100.00\% & 69.74\% & 100.00\% & 92.91\% \\
MFT & T5 & 75.62\% & 96.43\% & 62.06\% & 100.00\% & 70.16\% & 100.00\% & 91.97\% \\
MFT & T6 & 75.83\% & 96.55\% & 81.58\% & 100.00\% & 86.03\% & 100.00\% & 100.00\% \\
\cdashline{1-9}
DIR & T7 & 65.91\% & 71.93\% & 57.71\% & 73.68\% & 65.23\% & 71.93\% & 77.65\% \\
DIR & T8 & 36.23\% & 73.68\% & 66.40\% & 71.93\% & 75.44\% & 71.93\% & 99.99\% \\
DIR & T9 & 87.96\% & 64.91\% & 67.57\% & 64.91\% & 77.50\% & 61.40\% & 78.02\% \\
DIR & T10 & 61.31\% & 73.68\% & 60.74\% & 73.68\% & 67.37\% & 71.93\% & 81.65\% \\
DIR & T11 & 78.31\% & 77.19\% & 65.87\% & 80.70\% & 69.79\% & 77.19\% & 77.80\% \\
DIR & T12 & 80.98\% & 71.93\% & 66.64\% & 73.68\% & 67.08\% & 70.18\% & 77.86\% \\
DIR & T13 & 37.48\% & 73.68\% & 67.41\% & 68.42\% & 80.09\% & 71.93\% & 99.94\% \\
DIR & T14 & 77.87\% & 64.91\% & 62.64\% & 64.91\% & 69.95\% & 61.40\% & 77.99\% \\
DIR & T15 & 35.46\% & 73.68\% & 68.01\% & 73.68\% & 74.64\% & 71.93\% & 99.99\% \\
DIR & T16 & 88.37\% & 64.91\% & 67.57\% & 63.16\% & 74.96\% & 59.65\% & 78.06\% \\
DIR & T17 & 39.90\% & 73.68\% & 63.69\% & 73.68\% & 70.60\% & 71.93\% & 99.96\% \\
DIR & T18 & 86.83\% & 73.68\% & 49.92\% & 75.44\% & 77.18\% & 71.93\% & 78.14\% \\
DIR & T19 & 50.36\% & 64.91\% & 71.93\% & 61.40\% & 72.37\% & 61.40\% & 99.97\% \\
DIR & T20 & 77.34\% & 77.19\% & 58.76\% & 78.95\% & 65.35\% & 75.44\% & 82.81\% \\
DIR & T21 & 62.84\% & 73.68\% & 71.45\% & 73.68\% & 72.62\% & 71.93\% & 0.97\% \\
DIR & T22 & 37.24\% & 77.19\% & 60.90\% & 78.95\% & 73.02\% & 77.19\% & 78.11\% \\
DIR & T23 & 46.61\% & 64.91\% & 64.14\% & 59.65\% & 66.11\% & 61.40\% & 66.84\%
\end{tabular}
\end{adjustbox}
\caption{Accuracy rates of individual DECK tests for each model, fine-tuned on each dataset.}
\label{apptab:beh2}
\end{table*}

\end{document}